%% file: main.tex
\documentclass[accepted]{uai2021} 

\usepackage[american]{babel}
\usepackage{soul}
\usepackage{hyperref}
\usepackage{url}
\usepackage{amsmath}
\usepackage{amsthm}
\usepackage{amssymb}
\usepackage{multirow}
\usepackage{graphicx}
\usepackage{enumitem,kantlipsum}
\usepackage{caption}
\usepackage{subcaption}
\usepackage{xcolor}
\usepackage{natbib} 
\usepackage{mathtools} 
\usepackage{booktabs} 

\title{Batch Inverse-Variance Weighting: Deep Heteroscedastic Regression}

\author[1]{\href{mailto:Vincent Mai <vincent.mai@umontreal.ca>?Subject=Your BIV UAI 2021 paper}{Vincent~Mai}{}} 
\author[1]{Whaleed~Khamies}
\author[1,2]{Liam~Paull}

\affil[1]{%
    Robotics and Embodied AI Lab\\
    Mila - Quebec Institute of Artificial Intelligence\\
    Université de Montréal, Canada
}
\affil[2]{%
    Canada CIFAR AI Chair
}

\begin{document}
\maketitle

\begin{abstract}
Heteroscedastic regression is the task of supervised learning where each label is subject to noise from a different distribution. This noise can be caused by the labelling process, and impacts negatively the performance of the learning algorithm as it violates the i.i.d. assumptions. In many situations however, the labelling process is able to estimate the variance of such distribution for each label, which can be used as an additional information to mitigate this impact. We adapt an inverse-variance weighted mean square error, based on the Gauss-Markov theorem, for parameter optimization on neural networks. We introduce Batch Inverse-Variance, a loss function which is robust to near-ground truth samples, and allows to control the effective learning rate. Our experimental results show that BIV improves significantly the performance of the networks on two noisy datasets, compared to L2 loss, inverse-variance weighting, as well as a filtering-based baseline.
\end{abstract}

\input{Introduction}

\input{Background}

\input{Related_work}

\input{BIV}

\input{Exp_setup}

\input{Exp_results}

\input{Conclusion}

\begin{acknowledgements} 
    We acknowledge the support of the Natural Sciences and Engineering Research Council of Canada (NSERC) as well as the Canadian Institute for Advanced Research (CIFAR).

\end{acknowledgements}


\bibliography{Bibliography}

\pagebreak

\appendix
\section{Appendix - Data sets and Neural Networks}
\label{sec:app_datasets}
\input{A_DatasetsNN}
\section{Appendix - Additional experiments}
\input{B_AdditionalExperiments}
\end{document}

%% file: Introduction.tex
\section{Introduction}

In supervised learning, a central assumption is that the samples in the training dataset, used to train the model, and the samples in the testing set, used to evaluate the model, are sampled from identical distributions. Formally, for input $\mathbf{x}$ and label $y$, this assumption implies that $p_{\mathrm{train}}(\mathbf{x},y) = p_{\mathrm{test}}(\mathbf{x},y)$. This can be decomposed as the product $p_{\mathrm{train}}(\mathbf{x})\cdot p_{\mathrm{train}}(y|\mathbf{x})  = p_{\mathrm{test}}(\mathbf{x})\cdot p_{\mathrm{test}}(y|\mathbf{x}) $, which is true if : \begin{enumerate}
    \item The training dataset is representative and the features in both datasets are  sampled from the same distribution: $p_{\mathrm{train}}(\mathbf{x}) = p_{\mathrm{test}}(\mathbf{x})$.
    \item The labels in both datasets are sampled from the same conditional distribution: $p_{\mathrm{train}}(y|\mathbf{x}) = p_{\mathrm{test}}(y|\mathbf{x})$. If this condition is violated, the training labels are \textit{noisy}. 
\end{enumerate}

One case of noisy labels is when the labelling process induces uncertainty about the labels. In this case, $p_{\mathrm{train}}(y|\mathbf{x})$ encapsulates the uncertainty of the labelling process, whereas the ground truth $p_{\mathrm{test}}(y|\mathbf{x})$, over which we seek to optimize the performance of the model, are drawn from a Dirac delta distribution, even though it may be impossible to collect such a dataset in practice. In this case, the performance of the deployed model may decrease since the training process did not actually optimize the model's parameters based on the correct data \citep{Arpit_et_al_2017, Kawaguchi_Kaelbling_Bengio_2020}.

In this paper, we examine the case where we have some additional information about $p_{\mathrm{train}}(y|\mathbf{x})$. More specifically, we focus on the task of regression with a deep neural network when labels are corrupted by heteroscedastic noise: $p_{\mathrm{train}}(y|\mathbf{x})\sim \mathcal{N}(\mu_y,\sigma_y)$ where $\mu_y$ is the ground truth label and $\sigma_y$ is different for each sample. We assume that we have access to at least an estimate of the variance $\sigma_y$ for each corrupted each label. This information is available if the labels are being generated by some stochastic process that is capable of also reporting uncertainty. We examine how the knowledge of the estimate of the label noise variance can be used to mitigate the effect of the noise on the learning process of a deep neural network. We refer our method as Batch Inverse-Variance (BIV)\footnote{\href{https://github.com/montrealrobotics/BIV}{https://github.com/montrealrobotics/BIV}}, which is based on inverse-variance weighting for linear regression (IV). We show that applying the inverse-variance loss allows for a strong empirical advantage over the standard L2 loss. However, it is unstable in presence of near ground-truth labels and not appropriate for all optimizers. BIV includes two mechanisms to stabilize the learning process in this case.

Our claimed contributions are threefold:
\begin{enumerate}
    \item An analysis of the labelling process giving rise to a heteroscedastic regression problem in supervised learning.
    \item We present BIV, a loss function which uses the noise variance to minimize the performance loss due to the noise in a robust and reliable way while keeping control on the learning rate. 
    \item An experimental study comparing the performances of BIV over L2 loss, IV, as well as over a threshold-based filtering method.
\end{enumerate}

How best to use the information of the variance of heteroscedastically noisy labels in neural networks has not been studied in detail in the literature, maybe because neural networks are usually robust to strong label noise \citep{Rolnick_Veit_Belongie_Shavit_2018} or because the datasets traditionally do not include such information. We believe however that (1) the recent usage of deep neural networks in fields in which uncertainty plays a major role, such as robotics \citep{ProbaRobotics}, as well as (2) the combined advances in uncertainty estimation of the neural network output \citep{Kendall_Gal_2017, Lakshminarayanan_Pritzel_Blundell_2017} and in complex deep learning systems such as deep reinforcement learning where the labels are provided by another neural network \citep{DQN_2015, Haarnoja_Zhou_Abbeel_Levine_2018}, make this problem highly relevant.

\textbf{The outline of the paper is as follows:} In section 2, we describe the task of regression with heteroscedastic noisy labels. In section 3, we position our work among the existing literature on learning with noisy labels. In section 4, we explain the challenges of using IV in neural networks, and introduce BIV as an approach to improve its robustness. In section 5, we describe the setup for the experiments we made to validate the benefits of using BIV, and we present and analyze the results in section 6.

%% file: Background.tex
\section{Heteroscedastic noisy labels in regression}
\label{sec:generationHeteroNoisyLabels}

Consider an unlabelled dataset with inputs $\{\mathbf{x}_i\}$. To label it, one must apply to each input $\mathbf{x}_i$ an instance of a label generator $LG_j$ which should provide its associated true label $y_{i}$. This label generator has access to some features $\mathbf{z}_i \in \mathcal{Z}$ correlated to $\mathbf{x}_i$. We define $LG_j: \mathcal{Z} \xrightarrow[]{}\mathbb{R}$ .
When the labelling process is not exact and causes some noise on the label, the noisy label of $\mathbf{x}_i$ provided by $LG_j$ is defined as $\Tilde{y}_{i,j}$.
Noise on a measured or estimated value is often represented by a Gaussian distribution, based on the central limit theorem, as most noisy processes are the sum of several independent variables. Gaussian distributions are also mathematically practical, although they present some drawbacks as they can only represent uni-modal and symmetric noise. We model:
\begin{equation}
    \Tilde{y}_{i,j} = y_i + \delta_{y_{i,j}} \mathrm{\ with\ } \delta_{y_{i,j}} \sim N(0, \sigma^2_{i,j})
\end{equation}

$\sigma^2_{i,j}$ can be a function of $\mathbf{z}_i$ and $LG_j$, without any assumption on its dependence on one or the other. 
We finally assume that the label generator is able to provide an estimate of $\sigma^2_{i,j}$, therefore being re-defined as $LG_j: \mathcal{Z} \xrightarrow[]{}\mathbb{R}\times\mathbb{R}_{\geq 0}$. The training dataset is formed of triplets $(\mathbf{x}_i, \sigma^2_{i,j}, \Tilde{y}_{i,j})$, renamed $(\mathbf{x}_k, \sigma^2_k, \Tilde{y}_k)$ for triplet $k$ for simplicity. This setup describes many labelling processes, such as:

\paragraph{Crowd-sourced labelling:} In the example case of age estimation from facial pictures, labellers Alice and Bob are given $\mathbf{z}_i = \mathbf{x}_i$ the picture of someone's face and are asked to estimate the age of that person. Age is harder to estimate for older people come (5 and 15 years of age are harder to confuse than 75 and 85) suggesting a correlation between $\sigma^2_{i,j}$ and $\mathbf{z}_i$. But Alice and Bob may also have been given different instructions regarding the precision needed, inducing a correlation between $\sigma^2_{i,j}$ and $LG_j$. Finally, there may be some additional interactions between $\mathbf{z}_i$ and $LG_j$, as for example Alice may know Charlie, recognize him on the picture and label his age with lower uncertainty.
Both labellers can provide an estimation of the uncertainty around their labels, for example with a plus-minus range which can be used as a proxy for standard deviation.
    
\paragraph{Labelling from sensor readings, population studies, or simulations:} Imagine you want build a dataset of pictures $\mathbf{x}_i$ from a camera on the ground labelled with the position $y_i$ of a drone in the sky. To estimate the position of the drone at the moment the picture was taken, you could use state estimation algorithms based on the Bayes' filter \citep{ProbaRobotics}. These algorithms take as an input $\mathbf{z}_i$ the measurements from the drone's sensors, and provide a full posterior distribution over the state, sometimes under a Gaussian assumption for Kalman filters for example. The uncertainty depends on the precision of the sensors, the observability of a given state, the precision of the dynamic model, and the time since sensor signals were received. Similarly, studies based on population such as polling or pharmaceutical trials have quantified uncertainties based on the quantity and quality of their samples. It is also possible to train on simulators, as in climate sciences  \citep{Rasp_Pritchard_Gentine_2018} or in epidemiology \citep{Alsdurf_2020}, and some of them provide their estimations' uncertainty based on the simulation procedure and the inclusion of real measurements in the model.
    
\paragraph{Using predictions from a neural network in complex neural architectures:} In deep reinforcement learning for example, the critic network learns to predict a value from a state-action pair under the supervision of the heteroscedastic noisy output of a target network plus the reward \citep{DQN_2015, Haarnoja_Zhou_Abbeel_Levine_2018}. While the estimation of the uncertainty of the output of a neural network is not an easy task, it is an active field of research \citep{Gal_2016, Peretroukhin2019DeepPR}. There, $\mathbf{z}_i$ is the state-action pair at the next step, and $LG_j$ the target network being updated over time. The prediction is a mix of aleatoric and epistemic uncertainties as defined by \citet{Kendall_Gal_2017} which are dependent on both $\mathbf{z}_i$ and $LG_j$.

We could not find any current dataset that provides such label uncertainty information for regression. However, as it is precious information, we argue that it should actually be provided when possible. In classification, \cite{Xie2016CVPR, Kitti360} took a step in this direction by providing a ``confidence'' score from 0 to 255 for each pixel in the KITTI-360 dataset.

%% file: Related_work.tex
\section{Related work}
\label{sec:relatedwork}

Noise on labels amounts to a loss of information. When the noise is significant enough, it leads to overfitting and lower model performance \citep{832649,Zhang_Bengio_Hardt_Recht_Vinyals_2017}. This effect is more prevalent in small data settings \citep{Horn_2015}. 
Four possible strategies exist in the literature to tackle this problem for neural networks: detection, correction, robustness, or re-weighting.  
\textbf{Detection} consists of identifying noisy labels and ignoring them in the learning process. These methods are often based on the observation that neural networks first fit on consistent, non-noisy data \citep{Arpit_et_al_2017}, thus converging to a higher loss on the noisy samples \citep{Reed_Lee_Anguelov_Szegedy_Erhan_Rabinovich_2015,Shen_Sanghavi_2019}. Other methods use several neural networks to co-teach each other \citep{Han_Yao_Yu_Niu_Xu_Hu_Tsang_Sugiyama_2018, Yu_Han_Yao_Niu_Tsang_Sugiyama_2019} or dropout to estimate the consistency of the data \citep{Reed_Lee_Anguelov_Szegedy_Erhan_Rabinovich_2015}. However, in the case of imbalanced training datasets, higher loss can also be the signature of a non-noisy but rare sample. \citet{Cao_Chen_Lu_Arechiga_Gaidon_Ma_2020} address this  ambiguity by regularizing different regions of the input  space differently.
\textbf{Correction} strategies go further: once noise is detected, the noisy labels are changed to probability distributions. Such an operation requires a noise model. \citet{GoldbergerB17, pmlr_v84_kremer18a, Ma_Wang_Houle_Zhou_Erfani_Xia_Wijewickrema_Bailey_2018m, Tanno_Saeedi_Sankaranarayanan_Alexander_Silberman_2019, Yi_Wu_2019} learn it jointly with the parameters, assuming a correlation between the noise and the input, the labels, or both. 
\textbf{Robust loss functions} are less sensitive to noise. \citet{832649} proposed to avoid overfitting due to noise by ignoring samples during the training when the prediction error is reasonable. \citet{Natarajan_Dhillon_Ravikumar_Tewari_2013} compute the loss assuming knowledge of example-independent mislabelling probabilities in binary classification, and then optimize these hyperparameters with cross-validation. More recent works are based on reverse cross-entropy \citep{Wang_Ma_Chen_Luo_Yi_Bailey_2019} or curriculum loss \citep{Lyu_Tsang_2020}.  Others leverage a distillate of the information gathered from a subset of clean labels to guide training with noisy labels \citep{Li_Yang_Song_Cao_Luo_Li_2017}.
\textbf{Re-weighting} the samples is another efficient method for mitigating noise in datasets.  \citet{Liu_Tao_2016} estimate the effective label probabilities as well as noise rates for a given input and use these estimates to weigh the samples using importance sampling. \citet{Shu_Xie_Yi_Zhao_Zhou_Xu_Meng_2019} go one step further by learning the weighting function through a meta-learning  method.   \cite{Jenni_Favaro_2018} control overfitting by adjusting sample  weights in the training and validation mini-batches, increasing robustness to overfitting on noisy labels.

While most works that address noisy labels consider classification tasks \citep{Song_Kim_Park_Lee_2020}, only some of these strategies can be generalized to regression. Heteroscedastic regression occurs when each label's noise is sampled from a different distribution. \citet{Nix_1994} tackle this problem in neural networks by jointly training a variance estimator based on the maximum likelihood of an underlying Gaussian model. \citet{Kendall_Gal_2017} use the same idea to estimate the aleatoric (input-dependant) uncertainty of the network's prediction, while using dropout as a Bayesian approximation for the epistemic uncertainty (due to the learning process) as in \citep{Gal_2016}. \citet{Sai_Jinxia_Zhongxia_2009} use a re-weighting approach to regression, identifying noisy samples using the Parzen windowing method.

Similarly to \citet{Sai_Jinxia_Zhongxia_2009}, our method tackles heteroscedastic regression in neural networks using a re-weighting approach. The main distinction between our work and most of the related literature is that, while we do not require that the noise variance is a function of the input or of the label, we do assume that we have access to the noise variance, or at least an estimate of it. In addition, we do not seek to regress the variance of the model's prediction. This is significant compared to the previous works in both regression and classification as it proposes to answer the question: how to best use this additional information to train a deep neural network model?

Note that other regression models, such as Gaussian processes, have the ability to take the uncertainty of the label into account built into their framework. However, they require the design of a kernel to define distances between inputs, which is difficult in high dimensional inputs such as images. Additionally, they scale poorly with the number of samples \citep{Bishop_2006}.

%% file: BIV.tex
\section{Batch Inverse-Variance Weighting}
\subsection{Inverse variance for linear models}
\label{sec:HLR}

The task of heteroscedastic linear regression, where the model is linear, is solved by optimizing a weighted mean square error (WMSE) with inverse-variance weights, which is the optimal solution as per the Gauss-Markov theorem \citep{Shalizi}:
\begin{equation}
    WMSE = \sum_{k=0}^n \dfrac{(y_k - \mathbf{x_k}\cdot\beta)^2}{\sigma^2_k}
    \label{eq:HLR}
\end{equation}
where $\beta$ is the vector of parameters used as linear coefficients. This is also the solution to maximum likelihood estimation for parameters $\beta$ given the training examples \citep{Fisher_1957}.

While the solution to such an optimization is known for linear regression ($\beta^* = (\mathbf{x}^T\mathbf{wx}^{-1})\mathbf{x}^T \mathbf{wy}$), we could apply it to gradient-based methods for neural networks by using the Inverse Variance (IV) loss function for each sample:

\begin{equation}
    \mathcal{L}_{\mathrm{IV}}(\mathbf{x}_k,  \Tilde{y}_k, \theta) =  \dfrac{\left(f(\mathbf{x}_k, \theta) - \Tilde{y}_k\right)^2}{\sigma_k^2}
    \label{eq:IVLRweighted_loss}
\end{equation}

However, using $\mathcal{L}_{\mathrm{IV}}(\mathbf{x}_k,  \Tilde{y}_k, \theta)$ in a gradient descent setup brings two problems:
\begin{enumerate}
    \item Near ground-truth samples, with very small $\sigma_k^2$, have a disproportionate learning rate with respect to the others. They risk to cause overfitting while other samples are ignored. 
    \item  In gradient-based optimizers, the learning rate impacts the optimization process and should be controllable by the practitioner. In IV loss, the average of the sample weights $1/\sigma^2_k$ acts as a constant multiplied to the learning rate, making such control harder. While this would be less important in adaptive learning rate optimizers such as Adam \citep{Kingma_Ba_2017} as any constant multiplied to the loss is cancelled, it would still be problematic in continual or reinforcement learning tasks where the noise variance distribution can evolve. 
\end{enumerate}

\subsection{Batch Inverse-Variance weighting for heteroscedastic noisy labels}
\label{sec:BIV}

To address these issues, we introduce the Batch Inverse-Variance (BIV) loss function:
\begin{equation}
    \mathcal{L}_{\mathrm{BIV}}(D_i, \theta) = \left( \sum_{k=0}^K  \dfrac{1}{\sigma_k^2 + \epsilon}  \right)^{-1} \sum_{k=0}^K \dfrac{\mathcal{L}\left(f(\mathbf{x}_k, \theta),\Tilde{y}_k\right)}{\sigma_k^2 + \epsilon}
    \label{eq:IVweighted_loss}
\end{equation}

The first distinction with respect to the IV loss of (\ref{eq:IVLRweighted_loss}) is that it is computed on the whole batch instead of only being defined on a single sample. This allows to normalize the loss based on the weights of the other samples in the batch. The gradient’s scale is thus in dependent from the noise variance distribution, allowing better learning rate control and to use BIV in methods relying on the dynamics of the learning process.

Note that consistency is verified as, when $\sigma^2_k$ is identical for each sample, this formulation leads to empirical risk minimization normalized by the number of samples in the mini-batch.

Another difference with the IV loss is the smooth lower bound on the variance, $\epsilon$, which allows to incorporate samples with (near) ground-truth labels without ignoring the other samples. $\epsilon$ allows to control the effective batch size $EBS$, which, according to \cite{Kish_1965}, is equal to 
\begin{equation}
    EBS = \dfrac{\left(\sum_i^N w_i\right)^2}{\sum_i^N w_i^2} = \dfrac{\left(\sum_i^N \dfrac{1}{(\sigma_i^2 + \epsilon)}\right)^2}{\sum_i^N(\dfrac{1}{(\sigma_i^2 + \epsilon)})^2} 
\end{equation}
A higher $\epsilon$ increases $EBS$ by reducing the relative difference between the weights, thus reducing the effect of BIV. We found that $\epsilon$ can be set between $0.01$ and $0.1$ for normalized losses. More details can be found in appendix \ref{sec:epsilon}.

These two elements allow us to overcome the challenges related to inverse variance weighting applied to gradient descent described in section \ref{sec:HLR}.

\subsection{Cutoff: filtering heteroscedastic noisy labels }
\label{sec:cutoff_def}
In order to compare the BIV approach to a baseline, we introduce the Cutoff loss. As we do not assume that there is a correlation between $\mathbf{x}_k$ and $\sigma^2_k$, most correction and re-weighting algorithms as presented in Section \ref{sec:relatedwork} are not applicable. Additionally, most robust loss functions are specifically designed for classification problems. We thus compare BIV to a detection and rejection strategy.

In heteroscedastic regression, an important difference from classification with noisy labels is that all labels are corrupted, albeit not at the same scale. Defining which labels to ignore is therefore a matter of putting a threshold on the variance. As we have access to this information, detection strategies such as the ones used in section \ref{sec:relatedwork} are not necessary. Instead, we simply use an inverse Heaviside step function as a weight in the loss function:
\begin{equation}
    w_k = \mathbf{1}_{\sigma^2_k < C}
    \label{eq:cutoff_loss}
\end{equation}
where the threshold $C$ is a hyper-parameter. Similarly to equation (\ref{eq:IVweighted_loss}), we normalize the loss in the mini-batch by the sum of the weights, equal here to the number of samples considered as valid. As this filtering is equivalent to cutting off a part of the dataset, we refer to this method as `Cutoff', and consider it to be a relevant baseline to compare BIV against.

%% file: Exp_setup.tex
\section{Experimental Setup}
\label{sec:exp_setup}
To test the validity of the BIV loss (\ref{eq:IVweighted_loss}) approach, we compared its performance with the classical L2 loss as well as cutoff loss (\ref{eq:cutoff_loss}) on two datasets. We refer to ground-truth (GT) labels when training with L2 on noise-less data as the best performance that could be achieved on this dataset.

Unfortunately, we did not find any existing dataset for regression where label uncertainty is associated to the samples. We therefore used two UCI datasets \citep{UCI_Datasets} for regression cases, and artificially added noise to them. UTKFace Aligned\&Cropped \citep{UTKFace_2017} (UTKF), is a dataset for image-based age prediction. In the Bike Sharing dataset \citep{BikeSharing}, the task is to predict the number of bicycles rented in Washington D.C., from structured data containing the date, hour, and weather conditions. For UTKF, a convolutional neural network was used to predict the age, while a simple multi-layer perceptron was used for BikeSharing. More details about the datasets and models can be found in appendix \ref{sec:app_datasets}.

\subsection{Noise generation}
\label{sec:noise}
To produce the datasets $\{\mathbf{x}_k,\sigma^2_k, \Tilde{y}_k\}$
with noise as described in section \ref{sec:generationHeteroNoisyLabels}, we use a two-step process which does not assume any correlation between the noise and the state.
\begin{enumerate}
    \item The noise variance $\sigma^2_k$ is sampled from a distribution $P(\sigma^2)$ which only has support for $\sigma^2 \geq 0$
    \item $\Tilde{y}_k$ is sampled from a normal distribution $\mathcal{N}(y_k, \sigma^2_k)$.
\end{enumerate}

$P(\sigma^2)$ has a strong effect on the impact of BIV or Cutoff. For example, if it is a Dirac delta and all variances are the same, BIV becomes L2. We evaluate BIV on three different types of $P(\sigma^2)$. The average noise variance $\mu_P$ was chosen empirically so that the lowest test loss achieved by L2 is doubled compared to the ground-truth label case: $\mu_P = 2000$ for UTKF and $20000$ for BikeSharing.

\begin{itemize}[wide, labelwidth=0pt, labelindent=0pt]
\item[\textbf{Uniform distribution}]
The uniform distribution is characterized by its bounds $a,b$. Its expected value $\mu_P$ is the average of its bounds, and its variance $V = (b-a)^2/12$ . As $P$ only has support for $\sigma^2 \geq 0$, the maximum variance $V_{\mathrm{max}}$ is when $a=0$ and $b=2\mu_P$. While such a distribution  is not realistic, it is simple conceptually and allows for interesting insights.

\item[\textbf{``Binary uniform''}]
A more realistic distribution, which can also help us understand the effects of BIV, is when the data is generated from two regimes: low and high noise. We call the ``binary uniform'' distribution a mixture of two uniform distributions balanced by parameter $p$.

With probability $p$, the label is in a low noise regime: $\sigma^2 \sim U(0,1)$, with expected value $\mu_l = 0.5$. With probability $1-p$, the label is in a high noise regime: $\sigma^2 \sim U(a_h,b_h)$.
$a_h$ and $b_h$ are chosen such that the average is $\mu_{h}$ and the variance of the high-noise distribution is determined by $V_h \in [0, V_{\mathrm{max}}]$. Note that, if we want a given expected value of the whole distribution $\mu_P$, the value of $\mu_h$ changes depending on $p$: $\mu_h = (\mu_P-p\mu_l)/(1-p)$

Therefore, the high-noise expected value $\mu_h$ of the noise variance $\sigma^2$ in a distribution with high $p$ will be higher than the one for a low $p$, for the same value of $\mu_P$. In other words, a higher $p$ means more chance to be in the low-noise regime, but the high-noise regime is noisier.

\item[\textbf{Gamma distributions }] 
While the mixture of 2 uniform distributions ensures support in the low noise region, it is not continuous. We therefore also propose to use a Gamma distribution with shape parameter $\alpha$. If we want to control the expected value $\mu_P$, we adjust $\beta = \alpha/\mu_P$.

For a fixed expected value $\mu_P$, lower $\alpha$ and $\beta$ mean that there is a stronger support towards low variance noise, but the tail of the distribution spreads longer on the high-noise size. In other words, a lower $\alpha$ means more chances to have low-noise samples, but when they are noisy, the variance is higher. When $\alpha \leq 1$, the highest support is at $\sigma^2 = 0$.
\end{itemize}

\subsection{Evaluating the performance of the model}
The objective of BIV is to improve the performance at predicting the true label $y_i$. While a non-noisy test dataset may not be available in a real application, we aimed here at determining if BIV performs better than L2, and therefore measured the performance of the network using mean square error on ground-truth test data.

%% file: Exp_results.tex
\section{Experimental Results and Analysis}
\label{sec:results}

\subsection{For L2 loss, mean variance is all that matters}
\begin{figure}[h]
\begin{center}
\includegraphics[width=0.45\textwidth]{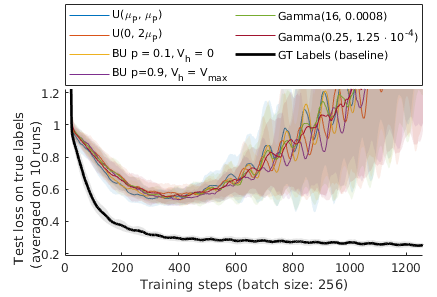}
\end{center}
\caption{Performance of the neural network on UTKF trained with L2 loss, for different $P(\sigma^2)$ with constant $\mu_P = 2000$.  No matter the distribution type or parameters, the performance is similar.} 
\label{fig:L2Var}
\end{figure}

Before looking at the results for BIV, we share an interesting insight for L2 loss with noisy labels which helps simplifying the analysis of the results.
Under the unbiased, heteroscedastic Gaussian-based noise model presented in section \ref{sec:noise}, the only parameter of distribution $P(\sigma)$ that mattered to describe the performance of the L2 loss is its average $\mu_P$, which is also the variance of the overall noise distribution. Independently of the distribution type, and the values of $V$, $p$ and $V_h$, or $\alpha$, as long as $\mu_P$ is equal, the L2 loss trained neural networks had the same performance.
This is shown in Figure \ref{fig:L2Var}. For the sake of clarity, all the curves in this section were smoothed using moving average with a 35 steps window, and the shaded area represents the standard deviation over several runs.

\subsection{High and low variance noise regimes: BIV acts as a filter}
\label{sec:res_binary}
With the binary uniform distribution, the noise is split in two regimes, with high or low variances. In this case, our results show that BIV performs better than L2, and actually similarly to the cutoff loss presented in section \ref{sec:cutoff_def} with a threshold $C = 1$.

\begin{figure}[h]
\begin{center}
\includegraphics[width=0.45\textwidth]{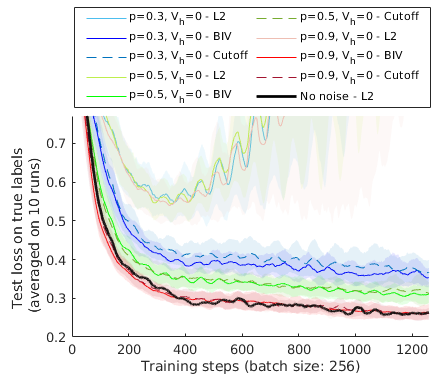}
\end{center}
\caption{Comparison between BIV, Cutoff and L2 losses for binary uniform distributions of variance with different $p$s for $\mu_P = 2000$ and $V_h=0$ on UTKF. }
\label{fig:BU_BIV_CO}
\end{figure}

Figure \ref{fig:BU_BIV_CO} compares the test losses on UTKF with different values of $p$ for $V_h=0$. While the L2 curves are strongly impacted by the noise, both the BIV and cutoff losses lead to better and very similar performances for a given $p$. When $p=0.3$, there are not a lot of information that can be used, and the performance is still impacted by the noise. When $p=0.9$, there is nearly as much near-ground-truth data as in the noiseless case, and the performance is comparable.

In the case of binary uniform distributions, BIV is acting as a filter, cutting off labels which are too noisy to contain any useful information.

\subsection{Continuous, decreasing noise variance distribution: the advantage of BIV}
\label{sec:res_gamma}
\begin{figure*}[h]
     \centering
     \includegraphics[width=\textwidth]{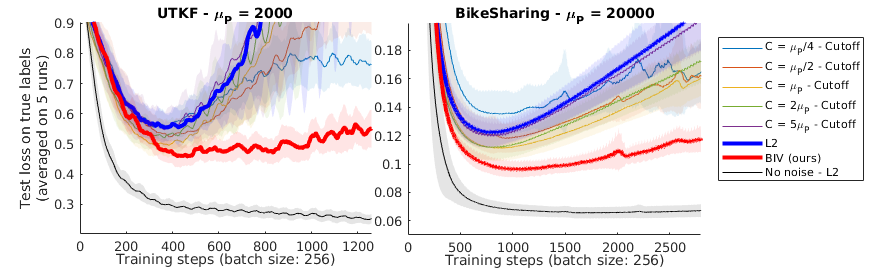}
    \caption{Comparison between the performances of BIV, L2, and different cutoff values on both datasets where the noise variance follows a Gamma distribution with $\alpha = 1$.}
    \label{fig:Gamma_BIV_CO}
\end{figure*}

On Gamma distributions, there is no clear threshold to define which information to use. When $\alpha \leq 1$, BIV shows a strong advantage compared to both L2 and cutoff. Figure \ref{fig:Gamma_BIV_CO} shows the results in both the BikeSharing and the UTKF datasets for Gamma distributions with $\alpha = 1$. 

In both datasets, when the cutoff parameter $C$ is too low ($\mu_P/4$ and $\mu_P/2$), there is not enough data to train the model. When $C$ is too high ($2 \mu_P$ and $5 \mu_P$), the data is too noisy and the curves go close to the original L2 loss. Even at the best case ($C = \mu_P$), cutoff is not better than BIV. This is because, in contrast to cutoff, BIV is able to extract some information from noisier samples while avoiding to over-fit on them.

In Table \ref{tab:GammaResults}, we present the lowest value of the test loss curves for the different methods with other $\alpha$ parameters for the Gamma distributions over both datasets. BIV consistently leads to the best performances, regardless of $P(\sigma^2)$. The plots showing these runs can be found in the appendix \ref{sec:app_gammaplots}. BIV is less sensitive to hyperparameters than cutoff, as it avoids the need to choose the right cutoff parameter for each distribution $P(\sigma^2)$. BIV's own hyperparameter $\epsilon$ can be set between 0.01 and 0.1 for any dataset with a normalized output, as shown in appendix \ref{sec:epsilon}, and be ready to use. As $\epsilon$ can be seen as a minimal variance, scaling it for other label distributions is straightforward: it suffices to multiply it by the variance of the label distribution. Here, it was set at 0.05.

\begin{table*}[h]
    \caption{Lowest test loss for different $\alpha$ on two datasets, for BIV, L2 and several cutoff losses. The test loss with standard deviation is computed as the average over 5 runs. In every case, BIV loss led to the lowest value. The best $C$ value differs based on $\alpha$.}
    \label{tab:GammaResults}
    \centering
    \begin{tabular}{|c|c|c|c|c|c|c|}
         \cline{2-7}
         \multicolumn{1}{c}{} & \multicolumn{2}{|c|}{$\alpha = 1$} & \multicolumn{2}{|c|}{$\alpha = 0.5$} & \multicolumn{2}{|c|}{$\alpha = 0.25$}  \\
         \cline{2-7}
         \multicolumn{1}{c|}{} & UTKF & Bike & UTKF & Bike & UTKF & Bike
         \\
         \hline
         $C = \mu_P/20$ & 0.79$\pm.08$ & 0.327$\pm.056$ & 0.48$\pm.04$ & 0.125$\pm.010$ & 0.38 $\pm.05$ & 0.092$\pm.008$ \\         \hline
         $C = \mu_P/4$  & 0.55$\pm.04$ & 0.135$\pm.016$ & 0.45$\pm.04$ & 0.097$\pm.006$ & 0.39$\pm.03$ & 0.085$\pm.006$ \\
          \hline
         $C = \mu_P$    & 0.50$\pm.04$ & 0.111$\pm.009$ & 0.48$\pm.04$ & 0.097$\pm.008$ & 0.43$\pm.03$ & 0.088$\pm.006$ \\
          \hline
         $C = 5\mu_P$   & 0.55$\pm.06$ & 0.120$\pm.009$ & 0.54$\pm.05$ & 0.111$\pm.012$ & 0.51$\pm.04$& 0.107$\pm.009$ \\
          \hline
         L2             & 0.56$\pm.05$ & 0.122$\pm.010$ & 0.56$\pm.05$ & 0.116$\pm.011$ & 0.55$\pm.05$ & 0.119 $\pm.012$ \\
         \hline
        BIV (ours)      & \textbf{0.46}$\pm.03$ & \textbf{0.096}$\pm.006$ & \textbf{0.40}$\pm.03$ & \textbf{0.088}$\pm.006$ & \textbf{0.33}$\pm.02$ & \textbf{0.079}$\pm.006$ \\
        \hline
        IV              & 0.99$\pm.03$ & 0.120$\pm.021$ & 1.65$\pm.03$ & 0.554$\pm.041$ & N.A. & N.A.\\
         \hline
         \hline
         GT labels      & 0.25$\pm.02$ & 0.066$\pm.004$ & 0.25$\pm.02$ & 0.066$\pm.004$ & 0.25$\pm.02$ & 0.066$\pm.004$ \\
         \hline
    \end{tabular}
\end{table*}

The benefit of BIV over L2 is clearly higher when $\alpha$ is lower. This is due to an increase in the support for low-variance noise in $P(\sigma^2)$. The more BIV can count on low-noise elements and differentiate them from high noise ones, the better it can perform. This is consistent with results from section \ref{sec:res_binary}, and with other experiments we have run. 
For example, when $\alpha > 1$, the highest support of $P$ is not at $\sigma^2 = 0$. BIV was less able to improve the performance compared to L2. 

We also ran the experiment with uniform distributions: the performance is better when variance $V$ is closer to $V_{max}$ (and $a$ to 0). But even when $V = V_{max}$, as there is less support in low noise variance than for Gamma distributions with $\alpha \leq 1$, the improvement is less important. 

In all cases, BIV was performing consistently better than L2 and at least better than cutoff in all the experiments we ran. More details on these results can be found in appendix \ref{sec:app_other_res}.

\subsection{Ablation study: comparison to IV}
\label{sec:ablation_study}
While we have shown in the previous parts that inverse variance weighting is indeed performing better than L2 or cutoff, we focus here on the role of both normalization and $\epsilon$ in BIV loss (\ref{eq:IVweighted_loss}) compared to a simple IV loss (\ref{eq:IVLRweighted_loss}) using an ablation study. The results of using IV on a Gamma distributions with $\alpha$ values of 1, 0.5 and 0.25 are shown in table \ref{tab:GammaResults}. It is clear that BIV performs better, but also that IV is struggling when $\alpha$ is small, which is when there is a higher probability that a sample's variance is very close to zero. Actually, the training process would diverge for $\alpha = 0.25$ on every seed for both BikeSharing and UTKF datasets.

\begin{figure*}[h]
     \centering
     \includegraphics[width=\textwidth]{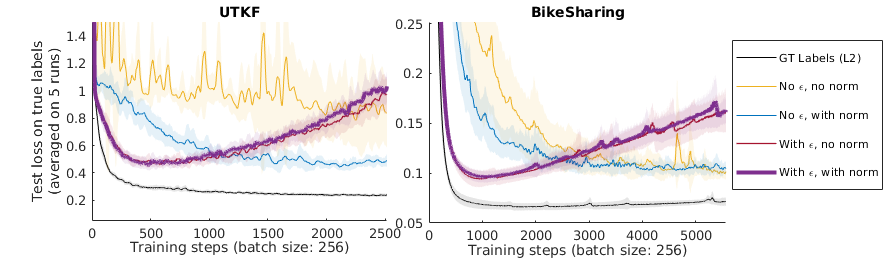}
    \caption{Ablation study for $\epsilon$ and normalization for both datasets where $P(\sigma^2)$ is a Gamma distribution with $\alpha = 1$.}
    \label{fig:AblationTest}
\end{figure*}

\begin{figure}[h]
     \centering
     \includegraphics[width=0.45\textwidth]{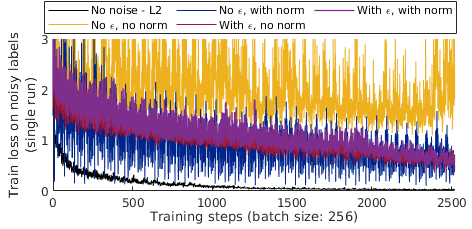}
    \caption{Training loss for ablation study for $\epsilon$ and normalization for both datasets where $P(\sigma^2)$ is a Gamma distribution with $\alpha = 1$. Note that, because the normalization value depends on $\epsilon$, these curves should not be compared quantitatively but quantitatively, by looking at their respective variability.}
    \label{fig:AblationTrain}
\end{figure}

More details can be seen in figure \ref{fig:AblationTest}. Here, the advantage of BIV, with $\epsilon$ and normalization, is clear over IV, without $\epsilon$ nor normalization. The effect of $\epsilon$ is significant. It induces not only the lowest test loss, but also learns faster by preventing low-variance samples to create high variability in the training loss. In figure \ref{fig:AblationTrain} showing the training loss, it is evident that with $\epsilon$, the variability of the loss is smaller. The impact of normalization is more subtle. In this case, the optimizer we used was Adam \citep{Kingma_Ba_2017}. Such optimizers are insensitive to a constant multiplied to the loss function, which is cancelled in the update process. Therefore, if the normalization constant is actually similar in every batch, it is equivalent to simply normalizing with the number of samples. This is what is happening when $\epsilon \neq 0$: the size of the minibatches and the limited variation in the noise variances makes this normalization close to effect-less. However, this is not the case when $\epsilon = 0$, where some mini-batches include such a small-variance sample that it makes the sum of the weights a lot higher than in other batches, which are actually ignored. With normalization, the effect of this near-ground truth sample is limited to the minibatch itself, which still allows for some learning when mini-batches are reshuffled and the samples have a chance to be taken into account at another epoch.

In the case of another optimizer such as stochastic gradient descent, the effect of normalization could be a lot more important, as it would also regulate the learning rate, regardless of the average sum of weights, and thus the distribution of noise variances in the dataset.

\subsection{Robustness}

We identified two factors that may impact the performances of BIV: the size of the mini-batches and the accuracy of the noise variance estimation. We tested the robustness of BIV when these factors are different than during the previous experiments.

\paragraph{Size of the mini-batches} In equation \ref{eq:IVweighted_loss}, each weight is normalized based on the assumption that the distribution of noise variances in the mini-batch is representative of the whole training dataset. With smaller mini-batches, the variability of the normalization constant between mini-batch should be bigger. However, this does not affect BIV, which still performs very well in these cases, as presented in section \ref{sec:Ap_Batchsize}. 

\paragraph{Noisy variances} Because the noise variance $\sigma_i^2$ is often estimated, the method needs to be robust to errors in $\sigma_i^2$'s. A model for the noise of $\sigma_i^2$ can be a Gaussian for which the variance is proportional to $\sigma_i^2$. In this case, results show that the effect of moderate to high levels of noise on BIV is not significant. More details can be seen in section \ref{sec:Ap_VarNoise}

%% file: Conclusion.tex
\section{Conclusion}
\label{sec:conclusion}

We have proposed Batch Inverse-Variance, a mini-batch based approach to account for the variance of the label noise in heteroscedastic regression tasks for neural networks. BIV is able to extract more information from the noisy dataset than L2 loss or threshold-based filtering approaches, and consistently outperforms them on both structured and unstructured datasets, by learning faster and reaching a lower test loss. The method is robust to low-variance, near ground-truth samples, which is not the case for naively applying an inverse-variance loss. It also enables reliable control of the learning rate, which is particularly important in tasks such as continual or reinforcement learning. BIV can be easily implemented in any regression setup, making it a versatile and ready to use algorithm.

An element which would need further investigation is the impact of such weights on the  performance of the model when the noise distribution is not uniformly distributed in the input space, to ensure that some regions of the input space are not doubly penalized by a higher level noise and a slower learning compared to the other regions.

%% file: A_DatasetsNN.tex
\subsection{UTKFace}

\subsubsection{Dataset description}

The UTKFace Aligned\&Cropped dataset \citep{UTKFace_2017} consists of 20,000 pictures of faces labelled with their age, ranging from 0 to 116 years. We use it in a regression setting: the network must predict the age of a person given the photo of their face. Unless described otherwise, 16,000 images were used for training, and 4,000 for testing. 

Some images are in black and white and some are in color. The pixel dimension of each image is 200x200.

Both the pixels and the labels were normalized before the training, so that their mean is 0 and standard deviation is 1 over the whole dataset. The noise variances were correspondingly scaled, as well as the cutoff threshold if applicable.

\subsubsection{Neural network and training hyper-parameters}
\label{sec:ApDetailsUTKF}

The model that we used was a Resnet-18 \citep{He_Zhang_Ren_Sun_2015}, not pretrained. It was trained with an Adam optimizer \citep{Kingma_Ba_2017}, a learning rate of 0.001 over 20 epochs. A batch size of 256 was used in order to ensure the best performance for the L2 method with noisy labels as well as to reduce the time necessary to the training process.

\subsection{Bike Sharing Dataset}

\subsubsection{Dataset description}

The Bike Sharing Dataset \citep{BikeSharing} consists of 17,379 samples of structured data. It contains, for nearly each hour of the years 2011 and 2012, the date, season, year, month, hour, day of the week, a boolean for it being a holiday, a boolean for it being a working day, the weather situation on a scale of 4 (1: clear and beautiful, 4: stormy or snowy), the temperature, the feeling temperature, the humidity, and the wind speed, in the city of Washington DC. It also contains the number of casual, registered, and total bike renters for each hour as recorded on the Capital Bikeshare system.

We use it in a regression setting: the network must predict the total number of bike renters given the time and weather information. Unless described otherwise, 7,000 samples were used for training, and 3,379 for testing. We used less samples than available for training because the low-data situation, noise has a stronger effect on the performance. The minimal test loss achieved with 7000 noiseless samples was very close to the one with 14000 samples, hinting that the additional samples did not give a lot of additional information.

We applied some pre-processing on the data to make it easier for the network to learn.
First, the date was normalized from a scale between day 1 to day 730 to a scale between 0 and $4\pi$. Then, we provided the network with the cosine and the sine of this number. This allowed to have the same representation for the same days of the year, while having the same distance between any two consecutive days, keeping the cyclic nature of a year. 
A similar idea was applied to hours, normalized from 0 to $2 \pi $ instead of 0 to 24, and with the cosine and sine given to the network. 
The day of the week, being a category, was given as a one-hot vector of dimension 7.
We also removed the season and the month as it was redundant information with the date.  

Overall, the number of features was 19: 
\begin{itemize}
    \item[1] Year
    \item[2-4] Date (sine and cos)
    \item[4-5] Hour (sine and cos)
    \item[6-12] Days of the week (one-hot vector)
    \item[13] Holiday boolean
    \item[14] Working day boolean
    \item[15] Weather situation
    \item[16] Temperature
    \item[17] Felt temperature
    \item[18] Humidity
    \item[19] Wind speed

\end{itemize}
We observed that the network was learning significantly faster and better provided with this format for the data.

Both the features and the labels were normalized before the training, so that their mean is 0 and standard deviation is 1 over the whole dataset. The noise variances were correspondingly scaled, as well as the cutoff threshold if applicable.

\subsubsection{Neural network and training hyper-parameters}

The model that we used was a multi-layer perceptron with 4 hidden layers, the first one with 100 neurons, then 50, 20, and 10. The activation function was ReLU. We did not use any additional technique such as batch normalization as it did not improve the performances.

The model was trained over 100 epochs on mini-batches of size 256 for similar reasons than explained in section \ref{sec:ApDetailsUTKF}, using the Adam optimizer with learning rate 0.001.

%% file: B_AdditionalExperiments.tex
\subsection{The influence of $\epsilon$}
\label{sec:epsilon}
In this section, we provide experimental results justifying our recommendation of the range of $[10^{-2}; 10^{-1}]$ for $\epsilon$. In the experiments presented in this article, we have mainly used  $\epsilon = 5\times10^{-2}$, and sometimes $\epsilon = 10^{-1}$.
$\epsilon$ must be chosen as part of a trade-off between mitigating the effect of BIV with near ground-truth labels while keeping its effect with noisy labels. To better understand these results, it is important to remember that $\epsilon$ is added to the variance that is used in the loss function. When the labels are normalized - which is the case in our work -, the noise and its variance for each label is normalized too. The value we recommend for $\epsilon$ should therefore be valid for any normalized set of labels.

The main role of $\epsilon$ is to set a maximal weight to the samples and prevent a near-zero variance sample to effectively reducing the minibatch to itself, thus ignoring the other potentially valid samples.
We tested several values of $\epsilon$ on both the UTKF and BikeSharing datasets, with $P(\sigma^2)$ being Gamma distributions with $\alpha = 1$ and $\mu_P = 2000$ and $\mu_P = 20000$ respectively.

\begin{figure}[!h]
\begin{center}
\includegraphics[width=0.45\textwidth]{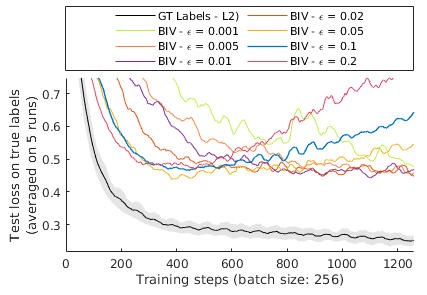}
\end{center}
\caption{Results of running BIV with different values of $\epsilon$ on UTKF with $P(\sigma^2)$ as a Gamma distribution with $\alpha = 1, \mu_P = 2000$.}
\label{fig:epsilonUTKF}
\end{figure}

\begin{figure}[!h]
\begin{center}
\includegraphics[width=0.45\textwidth]{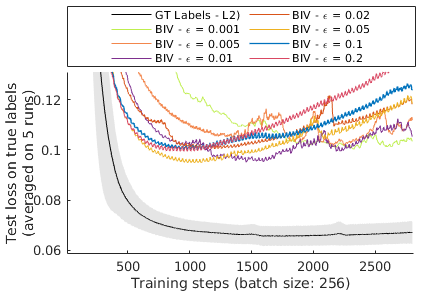}
\end{center}
\caption{Results of running BIV with different values of $\epsilon$ on BikeSharing with $P(\sigma^2)$ as a Gamma distribution with $\alpha = 1, \mu_P = 20000$.}
\label{fig:epsilonBike}
\end{figure}

The resulting graph can be seen in figures \ref{fig:epsilonUTKF} and \ref{fig:epsilonBike}

In both experiments, the optimal value for $\epsilon$ is 0.05. Between 0.02 and 0.1, the performances are acceptable. When $\epsilon$ is too small, the learning process gives too much importance to the low-noise samples, as seen in section \ref{sec:ablation_study}. When it is too high, it leans too much towards L2.

\begin{figure}[!h]
\begin{center}
\includegraphics[width=0.45\textwidth]{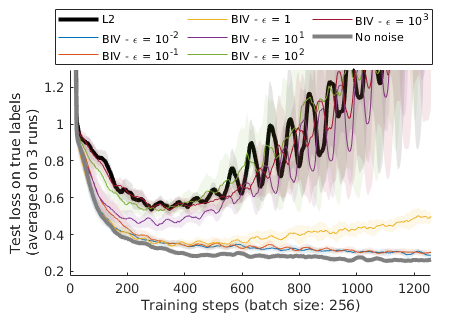}
\end{center}
\caption{Impact of $\epsilon$ when training with highly noisy labels using BIV loss on UTKF dataset. The variance was sampled through a binary uniform distribution with $p=0.5$, $\mu_P = 2000$, and $V_h = 0$. Very high $\epsilon$ shows a loss of performance as BIV approaches the L2 results. }
\label{fig:epsilonBinary}
\end{figure}

This can be seen in figure \ref{fig:epsilonBinary}, which applies BIV with different value for $\epsilon$ on a binary distribution on UTKF.

The fact that in the three, the same values for $\epsilon$ are optimal, even though tasks as well as the noise distribution are different, shows that these values are valid for most cases with normalized labels.

\subsection{BIV on different distributions}
\label{sec:app_other_res}

\subsubsection{Uniform distributions}
\begin{figure*}
    \centering
    \includegraphics[width=\textwidth]{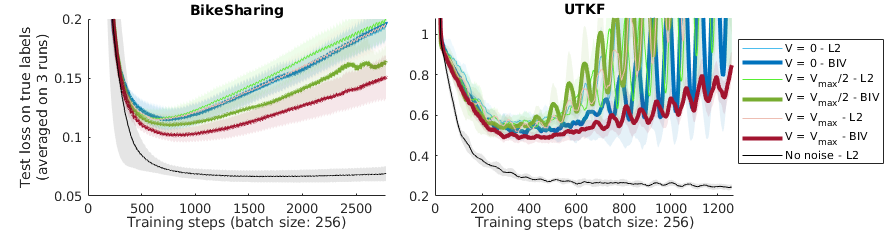}
    \caption{Test loss for L2 and BIV learning on uniform with different variances $V$.}
    \label{fig:Uniform_both}
\end{figure*}

We present in figure \ref{fig:Uniform_both} the results of the experiment with uniform distributions in more details. 

As explained in section \ref{sec:res_gamma}, we observe that BIV and L2 have the same performances when $V=0$ (and $a = b = \mu_P$). This is to be expected, as all samples have the exact same noise variance and thus the same weights. When $V = V_{max}$ ($a = 0$ and $b = 2\mu_P$), BIV has an advantage, as it is able to differentiate the samples and use the support of low-noise labels. When $V = V_{max}/2$ ($a = 0.293 \mu_P$ and $b = 1.707 \mu_P$), the difference between the samples is less important, and BIV only does a bit better than L2 on BikeSharing. On UTKF, the process has more variability and it is difficult to detect this effect.

In all cases, the benefit from using BIV is less important than with Gamma distributions with $\alpha \leq 1$, where the support on low-noise samples is higher.

\subsection{BIV on Gamma distributions}
\subsubsection{$\alpha \leq 1$}
\label{sec:app_gammaplots}
\begin{figure*}
    \centering
    \includegraphics[width=\textwidth]{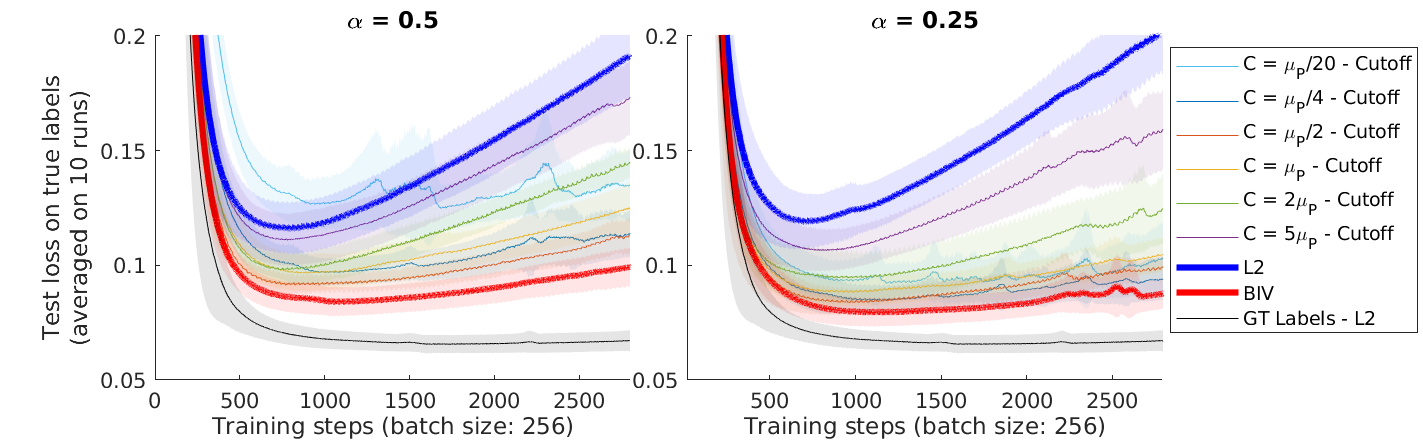}
    \caption{Test loss on the BikeSharing dataset, with $\alpha \leq 1$}
    \label{fig:GammaBikeLowAlpha}
\end{figure*}

\begin{figure*}
    \centering
    \includegraphics[width=\textwidth]{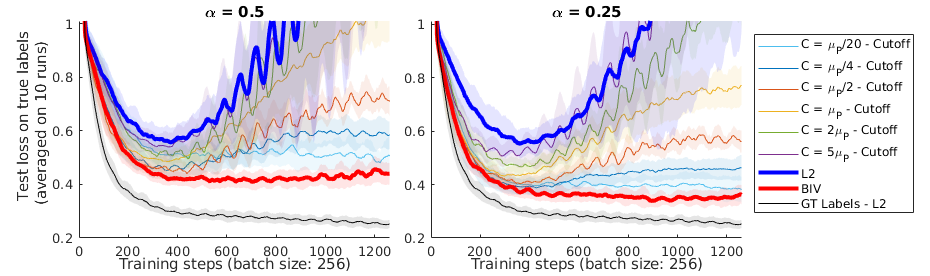}
    \caption{Test loss on the UTKF dataset, with $\alpha \leq 1$ }
    \label{fig:GammaUTKFLowAlpha}
\end{figure*}

As described in section \ref{sec:res_gamma}, the smaller $\alpha$, the better the performance of BIV and cutoffs. We show in figures \ref{fig:GammaBikeLowAlpha} and \ref{fig:GammaUTKFLowAlpha} the curves that led to the numbers in Table \ref{tab:GammaResults}. BIV consistently outperforms the other methods. The performance of cutoff methods strongly depends on $C$, and the best value of $C$ is not the same for every distribution $P$.

\begin{figure}[h]
    \centering
    \includegraphics[width=0.42\textwidth]{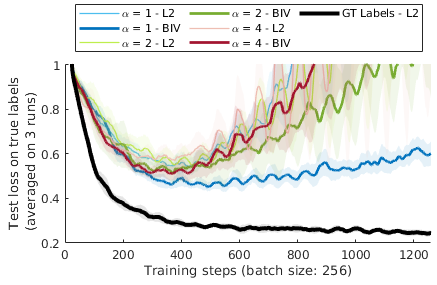}
    \caption{Test loss on the UTKF dataset for L2 and BIV learning on Gamma function with $\alpha \geq 1$.}
    \label{fig:GammaHigh}
\end{figure}

\subsubsection{$\alpha > 1$}
When $\alpha > 1$, the highest support of $P(\sigma^2)$ shifts towards $\mu_P$. This makes the samples less distinguishable for BIV and therefore the benefits of using it are reduced. This is shown in Figure \ref{fig:GammaHigh} on UTKF.

\subsection{Robustness of BIV}
\subsubsection{Size of the mini-batches}
\label{sec:Ap_Batchsize}

In equation (\ref{eq:IVweighted_loss}), the normalization constant is computed from the samples in the mini-batch. If the distribution of the noise variances in mini-batch is representative of the whole training dataset, the relative weight given to each sample with respect to the others is the same than if the normalization was made over the whole dataset. The larger the mini-batch, the more representative it is. In our experiments, we used a size of 256, which is arguably high. We tested our algorithm with lower batch sizes, from 16 to 128, to see if it was a critical factor in the performances.

\begin{figure*}
    \centering
    \includegraphics[width=\textwidth]{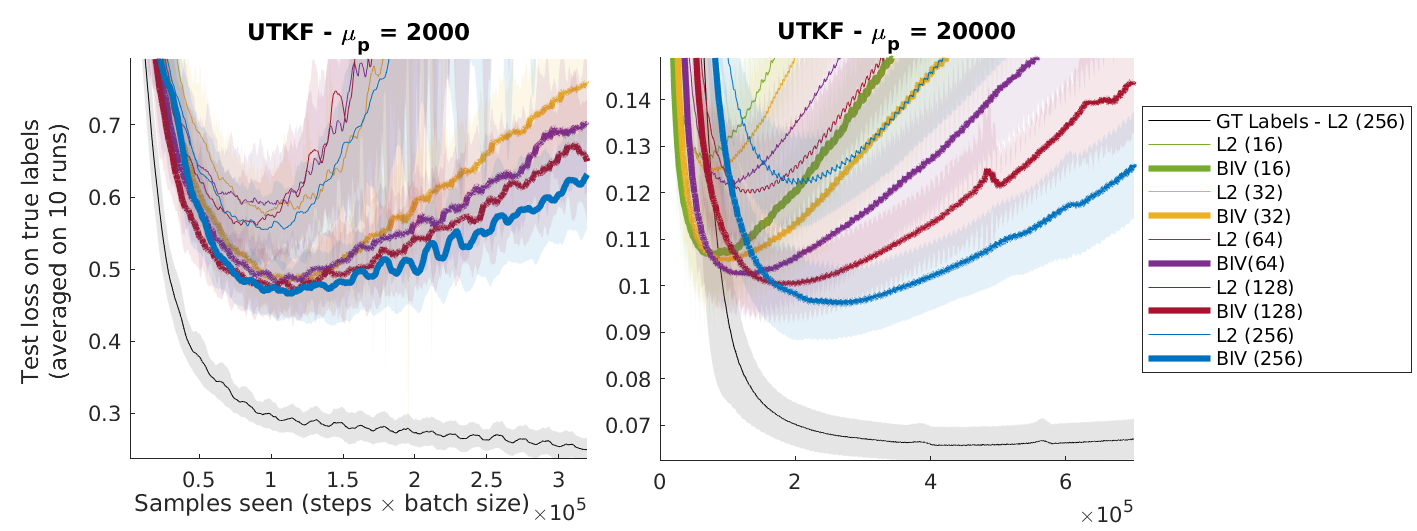}
    \caption{BIV with different batch sizes in both UTKF and BikeSharing datasets.}
    \label{fig:batchSize}
\end{figure*}

The results are presented in figure \ref{fig:batchSize}. In UTKF, the batch size does not make any significant difference in the performance with respect to the amount of samples seen, except for a slightly steeper overfitting once the best loss has been achieved. In BikeSharing, a smaller batch size makes the training faster with respect to the amount of samples, but with a higher minimal loss, for both L2 and BIV. While a larger batch size leads to a lower loss function, the effect of BIV compared to the corresponding L2 curve is not compromised by smaller batch-sizes.

Two main factors may explain this robustness. First, a mini-batch of size 16 seems to be already representative enough of the whole dataset for the purpose of normalization. Second, the fact that the mini-batches are populated differently at every epoch improves the robustness as a sample who would have been in a non-representative batch at one epoch may not be at another epoch. In any case, the size of the mini-batch is not a critical factor for BIV.

\subsubsection{Noisy variance estimation}
\label{sec:Ap_VarNoise}
In many scenarios, the variance $\sigma^2$ from which the noise was sampled is estimated, or inferred from a proxy, and therefore prone to be noisy itself. We tested the robustness of our method to such variance noise. In this experimental setup, the value given to the BIV algorithm is disturbed by noise $\delta_{\sigma_i^2}$.
We modelled this noise on $\sigma_i^2$ to be sampled from a normal distribution whose standard deviation is proportional to $\sigma_i^2$ with a coefficient of variance disturbance $D_v$:
\begin{equation}
    \delta_{\sigma_i^2} \sim \mathcal{N}(0, D_v \sigma_i^2/9)
\end{equation}

Dividing $\sigma_i^2$ by 9 allows to scale $D_v$ so that, when $D_v = 1$, $\delta_{\sigma_i^2} < -\sigma_i^2$ is at 3 standard deviations from the mean.

We then compute the noisy variance, which needs to be positive, as $\Tilde{\sigma}_i^2 = \left|\sigma_i^2 + \delta_{\sigma_i^2}\right|$.
The noise is therefore biased, but when $D_v \leq 1$, this is negligible as it happens with probability less than 0.15\%.

\begin{figure*}
    \centering
    \includegraphics[width=\textwidth]{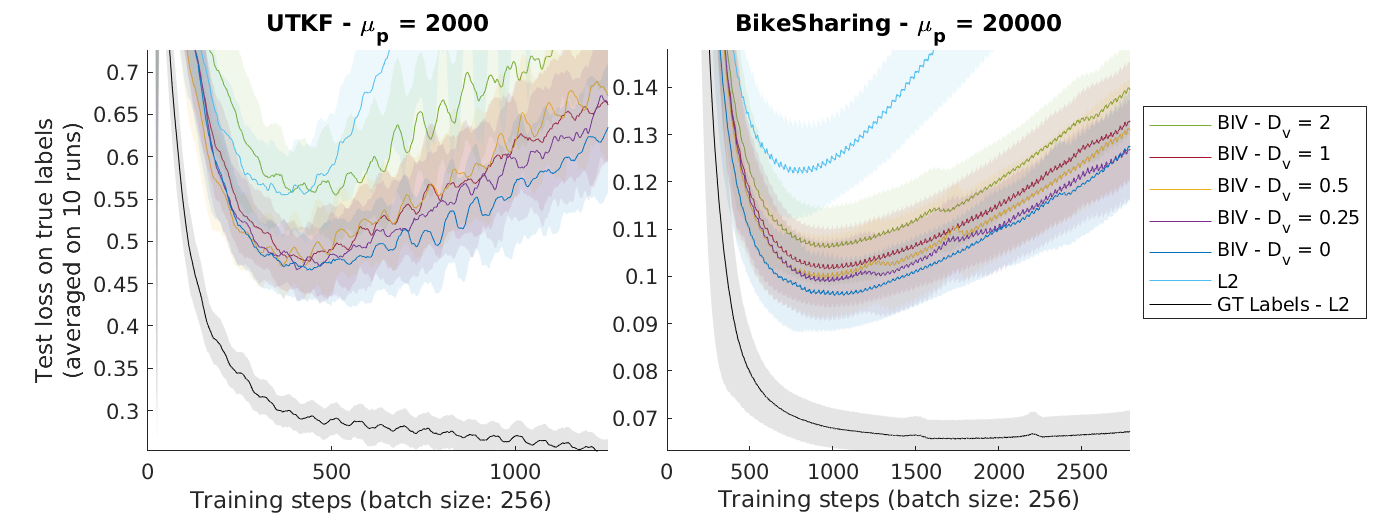}
    \caption{Robustness of BIV to noise in the variance with different disturbance coefficients $D_v$.}
    \label{fig:VarNoise}
\end{figure*}

The results presented in figure \ref{fig:VarNoise} show that, when $D_v \leq 1$, BIV is robust to such noise. While a higher $D_v$ leads to lower performance, the impact is small compared to the effect of BIV. However, when $D_v =2$, which is an arguably high level of noise and leads to bias as explained previously, the beneficial effect of BIV is significantly affected in BikeSharing, and completely disappears in UTKF.

\begin{figure*}
    \centering
    \includegraphics[width=\textwidth]{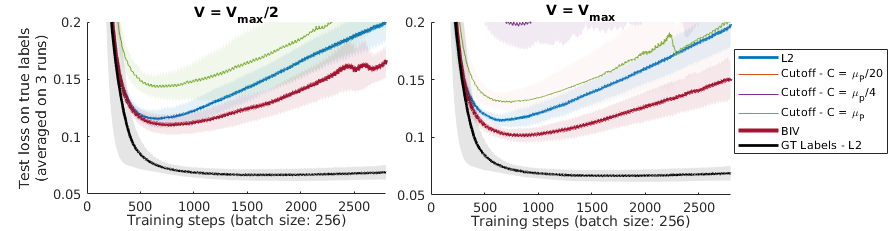}
    \caption{On BikeSharing with $\mu_P = 20000$, using cutoff is not helpful in the uniform setting. This is due to the significant loss of information induced by such a strategy.}
    \label{fig:UniformBikeCutoff}
\end{figure*}

In this setting, we also show as shown in figure \ref{fig:UniformBikeCutoff} that cutting off the noisy data is not a good strategy, as it always performs worse than L2.

%% file: main.bbl
\begin{thebibliography}{46}
\providecommand{\natexlab}[1]{#1}
\providecommand{\url}[1]{\texttt{#1}}
\expandafter\ifx\csname urlstyle\endcsname\relax
  \providecommand{\doi}[1]{doi: #1}\else
  \providecommand{\doi}{doi: \begingroup \urlstyle{rm}\Url}\fi

\bibitem[Alsdurf et~al.(2020)Alsdurf, Bengio, Deleu, Gupta, Ippolito, Janda,
  Jarvie, Kolody, Krastev, Maharaj, and et~al.]{Alsdurf_2020}
Hannah Alsdurf, Yoshua Bengio, Tristan Deleu, Prateek Gupta, Daphne Ippolito,
  Richard Janda, Max Jarvie, Tyler Kolody, Sekoul Krastev, Tegan Maharaj, and
  et~al.
\newblock Covi white paper.
\newblock \emph{arXiv:2005.08502 [cs]}, May 2020.

\bibitem[Arpit et~al.(2017)Arpit, Jastrzębski, Ballas, Krueger, Bengio,
  Kanwal, Maharaj, Fischer, Courville, Bengio, and et~al.]{Arpit_et_al_2017}
Devansh Arpit, Stanisław Jastrzębski, Nicolas Ballas, David Krueger, Emmanuel
  Bengio, Maxinder~S. Kanwal, Tegan Maharaj, Asja Fischer, Aaron Courville,
  Yoshua Bengio, and et~al.
\newblock A closer look at memorization in deep networks.
\newblock \emph{arXiv:1706.05394 [cs, stat]}, Jul 2017.

\bibitem[Bishop(2006)]{Bishop_2006}
Christopher~M. Bishop.
\newblock \emph{Pattern recognition and machine learning}.
\newblock Information science and statistics. Springer, 2006.
\newblock ISBN 978-0-387-31073-2.

\bibitem[Cao et~al.(2020)Cao, Chen, Lu, Arechiga, Gaidon, and
  Ma]{Cao_Chen_Lu_Arechiga_Gaidon_Ma_2020}
Kaidi Cao, Yining Chen, Junwei Lu, Nikos Arechiga, Adrien Gaidon, and Tengyu
  Ma.
\newblock Heteroskedastic and imbalanced deep learning with adaptive
  regularization.
\newblock \emph{arXiv:2006.15766 [cs, stat]}, Jun 2020.

\bibitem[Dua and Graff(2017)]{UCI_Datasets}
Dheeru Dua and Casey Graff.
\newblock {UCI} machine learning repository, 2017.
\newblock URL \url{http://archive.ics.uci.edu/ml}.

\bibitem[Fanaee-T and Gama(2013)]{BikeSharing}
Hadi Fanaee-T and Joao Gama.
\newblock Event labeling combining ensemble detectors and background knowledge.
\newblock \emph{Progress in Artificial Intelligence}, pages 1--15, 2013.

\bibitem[Fisher(1957)]{Fisher_1957}
G.~R. Fisher.
\newblock Maximum likelihood estimators with heteroscedastic errors.
\newblock \emph{Review of the International Statistical Institute}, 25\penalty0
  (1/3):\penalty0 52, 1957.

\bibitem[Gal and Ghahramani(2016)]{Gal_2016}
Yarin Gal and Zoubin Ghahramani.
\newblock Dropout as a bayesian approximation: Representing model uncertainty
  in deep learning.
\newblock In \emph{Proceedings of the 33rd International Conference on Machine
  Learning}, volume~48, page 1050–1059, 2016.

\bibitem[Goldberger and Ben{-}Reuven(2017)]{GoldbergerB17}
Jacob Goldberger and Ehud Ben{-}Reuven.
\newblock Training deep neural-networks using a noise adaptation layer.
\newblock In \emph{5th International Conference on Learning Representations},
  2017.

\bibitem[Haarnoja et~al.(2018)Haarnoja, Zhou, Abbeel, and
  Levine]{Haarnoja_Zhou_Abbeel_Levine_2018}
Tuomas Haarnoja, Aurick Zhou, Pieter Abbeel, and Sergey Levine.
\newblock Soft actor-critic: Off-policy maximum entropy deep reinforcement
  learning with a stochastic actor.
\newblock \emph{arXiv:1801.01290 [cs, stat]}, Aug 2018.
\newblock URL \url{http://arxiv.org/abs/1801.01290}.

\bibitem[{Han} et~al.(2018){Han}, {Yao}, {Yu}, {Niu}, {Xu}, {Hu}, {Tsang}, and
  {Sugiyama}]{Han_Yao_Yu_Niu_Xu_Hu_Tsang_Sugiyama_2018}
Bo~{Han}, Quanming {Yao}, Xingrui {Yu}, Gang {Niu}, Miao {Xu}, Weihua {Hu},
  Ivor {Tsang}, and Masashi {Sugiyama}.
\newblock {Co-teaching: Robust Training of Deep Neural Networks with Extremely
  Noisy Labels}.
\newblock \emph{arXiv e-prints}, art. arXiv:1804.06872, April 2018.

\bibitem[He et~al.(2015)He, Zhang, Ren, and Sun]{He_Zhang_Ren_Sun_2015}
Kaiming He, Xiangyu Zhang, Shaoqing Ren, and Jian Sun.
\newblock Deep residual learning for image recognition.
\newblock \emph{arXiv:1512.03385 [cs]}, Dec 2015.

\bibitem[Jenni and Favaro(2018)]{Jenni_Favaro_2018}
Simon Jenni and Paolo Favaro.
\newblock Deep bilevel learning.
\newblock \emph{arXiv:1809.01465 [cs, stat]}, Sep 2018.

\bibitem[Kawaguchi et~al.(2020)Kawaguchi, Kaelbling, and
  Bengio]{Kawaguchi_Kaelbling_Bengio_2020}
Kenji Kawaguchi, Leslie~Pack Kaelbling, and Yoshua Bengio.
\newblock Generalization in deep learning.
\newblock \emph{arXiv:1710.05468 [cs, stat]}, Jul 2020.

\bibitem[Kendall and Gal(2017)]{Kendall_Gal_2017}
Alex Kendall and Yarin Gal.
\newblock What uncertainties do we need in bayesian deep learning for computer
  vision?
\newblock In \emph{Advances in Neural Information Processing Systems 30}, pages
  5574--5584. Curran Associates, Inc., 2017.

\bibitem[Kingma and Ba(2017)]{Kingma_Ba_2017}
Diederik~P. Kingma and Jimmy Ba.
\newblock Adam: A method for stochastic optimization.
\newblock \emph{arXiv:1412.6980 [cs]}, Jan 2017.

\bibitem[Kish(1965)]{Kish_1965}
L.~Kish.
\newblock \emph{Survey Sampling}.
\newblock Wiley, 1965.
\newblock ISBN 978-0-471-48900-9.

\bibitem[Kremer et~al.(2018)Kremer, Sha, and Igel]{pmlr_v84_kremer18a}
Jan Kremer, Fei Sha, and Christian Igel.
\newblock Robust active label correction.
\newblock volume~84 of \emph{Proceedings of Machine Learning Research}, pages
  308--316, Playa Blanca, Lanzarote, Canary Islands, 09--11 Apr 2018. PMLR.
\newblock URL \url{http://proceedings.mlr.press/v84/kremer18a.html}.

\bibitem[Lakshminarayanan et~al.(2017)Lakshminarayanan, Pritzel, and
  Blundell]{Lakshminarayanan_Pritzel_Blundell_2017}
Balaji Lakshminarayanan, Alexander Pritzel, and Charles Blundell.
\newblock Simple and scalable predictive uncertainty estimation using deep
  ensembles.
\newblock \emph{arXiv:1612.01474 [cs, stat]}, Nov 2017.
\newblock URL \url{http://arxiv.org/abs/1612.01474}.
\newblock arXiv: 1612.01474.

\bibitem[Li et~al.(2017)Li, Yang, Song, Cao, Luo, and
  Li]{Li_Yang_Song_Cao_Luo_Li_2017}
Yuncheng Li, Jianchao Yang, Yale Song, Liangliang Cao, Jiebo Luo, and Li-Jia
  Li.
\newblock Learning from noisy labels with distillation.
\newblock \emph{arXiv:1703.02391 [cs, stat]}, Apr 2017.

\bibitem[Liu and Tao(2016)]{Liu_Tao_2016}
Tongliang Liu and Dacheng Tao.
\newblock Classification with noisy labels by importance reweighting.
\newblock \emph{IEEE Transactions on Pattern Analysis and Machine
  Intelligence}, 38\penalty0 (3):\penalty0 447–461, Mar 2016.
\newblock arXiv: 1411.7718.

\bibitem[{Liu} and {Castagna}(1999)]{832649}
Z.~P. {Liu} and J.~P. {Castagna}.
\newblock Avoiding overfitting caused by noise using a uniform training mode.
\newblock In \emph{IJCNN'99. International Joint Conference on Neural Networks.
  Proceedings (Cat. No.99CH36339)}, volume~3, pages 1788--1793 vol.3, 1999.

\bibitem[Lyu and Tsang(2020)]{Lyu_Tsang_2020}
Yueming Lyu and Ivor~W. Tsang.
\newblock Curriculum loss: Robust learning and generalization against label
  corruption.
\newblock \emph{arXiv:1905.10045 [cs, stat]}, Feb 2020.

\bibitem[Ma et~al.(2018)Ma, Wang, Houle, Zhou, Erfani, Xia, Wijewickrema, and
  Bailey]{Ma_Wang_Houle_Zhou_Erfani_Xia_Wijewickrema_Bailey_2018m}
Xingjun Ma, Yisen Wang, Michael~E. Houle, Shuo Zhou, Sarah~M. Erfani, Shu-Tao
  Xia, Sudanthi Wijewickrema, and James Bailey.
\newblock Dimensionality-driven learning with noisy labels.
\newblock \emph{arXiv:1806.02612 [cs, stat]}, Jul 2018.

\bibitem[Mnih et~al.(2015)Mnih, Kavukcuoglu, Silver, Rusu, Veness, Bellemare,
  Graves, Riedmiller, Fidjeland, Ostrovski, and et~al.]{DQN_2015}
Volodymyr Mnih, Koray Kavukcuoglu, David Silver, Andrei~A. Rusu, Joel Veness,
  Marc~G. Bellemare, Alex Graves, Martin Riedmiller, Andreas~K. Fidjeland,
  Georg Ostrovski, and et~al.
\newblock Human-level control through deep reinforcement learning.
\newblock \emph{Nature}, 518\penalty0 (7540):\penalty0 529–533, Feb 2015.

\bibitem[Natarajan et~al.(2013)Natarajan, Dhillon, Ravikumar, and
  Tewari]{Natarajan_Dhillon_Ravikumar_Tewari_2013}
Nagarajan Natarajan, Inderjit~S Dhillon, Pradeep~K Ravikumar, and Ambuj Tewari.
\newblock Learning with noisy labels.
\newblock In C.~J.~C. Burges, L.~Bottou, M.~Welling, Z.~Ghahramani, and K.~Q.
  Weinberger, editors, \emph{Advances in Neural Information Processing
  Systems}, volume~26, page 1196–1204. Curran Associates, Inc., 2013.

\bibitem[{Nix} and {Weigend}(1994)]{Nix_1994}
D.~A. {Nix} and A.~S. {Weigend}.
\newblock Estimating the mean and variance of the target probability
  distribution.
\newblock In \emph{Proceedings of 1994 IEEE International Conference on Neural
  Networks (ICNN'94)}, volume~1, pages 55--60 vol.1, 1994.

\bibitem[Peretroukhin et~al.(2019)Peretroukhin, Wagstaff, and
  Kelly]{Peretroukhin2019DeepPR}
Valentin Peretroukhin, Brandon Wagstaff, and Jonathan Kelly.
\newblock Deep probabilistic regression of elements of so(3) using quaternion
  averaging and uncertainty injection.
\newblock In \emph{CVPR Workshops}, 2019.

\bibitem[Rasp et~al.(2018)Rasp, Pritchard, and
  Gentine]{Rasp_Pritchard_Gentine_2018}
Stephan Rasp, Michael~S. Pritchard, and Pierre Gentine.
\newblock Deep learning to represent subgrid processes in climate models.
\newblock \emph{Proceedings of the National Academy of Sciences}, 115\penalty0
  (39):\penalty0 9684–9689, Sep 2018.

\bibitem[Reed et~al.(2015)Reed, Lee, Anguelov, Szegedy, Erhan, and
  Rabinovich]{Reed_Lee_Anguelov_Szegedy_Erhan_Rabinovich_2015}
Scott Reed, Honglak Lee, Dragomir Anguelov, Christian Szegedy, Dumitru Erhan,
  and Andrew Rabinovich.
\newblock Training deep neural networks on noisy labels with bootstrapping.
\newblock \emph{arXiv:1412.6596 [cs]}, Apr 2015.

\bibitem[Rolnick et~al.(2018)Rolnick, Veit, Belongie, and
  Shavit]{Rolnick_Veit_Belongie_Shavit_2018}
David Rolnick, Andreas Veit, Serge Belongie, and Nir Shavit.
\newblock Deep learning is robust to massive label noise.
\newblock \emph{arXiv:1705.10694 [cs]}, Feb 2018.

\bibitem[Sai et~al.(2009)Sai, Jinxia, and Zhongxia]{Sai_Jinxia_Zhongxia_2009}
Y.~Sai, R.~Jinxia, and L.~Zhongxia.
\newblock Learning of neural networks based on weighted mean squares error
  function.
\newblock In \emph{2009 Second International Symposium on Computational
  Intelligence and Design}, volume~1, page 241–244, Dec 2009.
\newblock \doi{10.1109/ISCID.2009.67}.

\bibitem[Shalizi(2019)]{Shalizi}
Cosma~Rohilla Shalizi.
\newblock \emph{Advanced Data Analysis from an Elementary Point of View}, 2019.
\newblock URL \url{https://www.stat.cmu.edu/~cshalizi/ADAfaEPoV/ADAfaEPoV.pdf}.
\newblock (Accessed November 13th, 2020).

\bibitem[Shen and Sanghavi(2019)]{Shen_Sanghavi_2019}
Yanyao Shen and Sujay Sanghavi.
\newblock Learning with bad training data via iterative trimmed loss
  minimization.
\newblock \emph{arXiv:1810.11874 [cs, stat]}, Feb 2019.

\bibitem[Shu et~al.(2019)Shu, Xie, Yi, Zhao, Zhou, Xu, and
  Meng]{Shu_Xie_Yi_Zhao_Zhou_Xu_Meng_2019}
Jun Shu, Qi~Xie, Lixuan Yi, Qian Zhao, Sanping Zhou, Zongben Xu, and Deyu Meng.
\newblock Meta-weight-net: Learning an explicit mapping for sample weighting.
\newblock \emph{arXiv:1902.07379 [cs, stat]}, Sep 2019.

\bibitem[Song et~al.(2020)Song, Kim, Park, and Lee]{Song_Kim_Park_Lee_2020}
Hwanjun Song, Minseok Kim, Dongmin Park, and Jae-Gil Lee.
\newblock Learning from noisy labels with deep neural networks: A survey.
\newblock \emph{arXiv:2007.08199 [cs, stat]}, Jul 2020.

\bibitem[Song and Zhang(2017)]{UTKFace_2017}
Yang Song and Zhifei Zhang.
\newblock \emph{UTKFace, Large Scale Face Dataset}, 2017.
\newblock URL \url{https://susanqq.github.io/UTKFace/}.
\newblock (Accessed June 11, 2020).

\bibitem[Tanno et~al.(2019)Tanno, Saeedi, Sankaranarayanan, Alexander, and
  Silberman]{Tanno_Saeedi_Sankaranarayanan_Alexander_Silberman_2019}
Ryutaro Tanno, Ardavan Saeedi, Swami Sankaranarayanan, Daniel~C. Alexander, and
  Nathan Silberman.
\newblock Learning from noisy labels by regularized estimation of annotator
  confusion.
\newblock In \emph{2019 IEEE/CVF Conference on Computer Vision and Pattern
  Recognition (CVPR)}, page 11236–11245. IEEE, Jun 2019.

\bibitem[Thrun et~al.(2006)Thrun, Burgard, and Fox]{ProbaRobotics}
Sebastian Thrun, Wolfram Burgard, and Dieter Fox.
\newblock \emph{Probabilistic Robotics}.
\newblock Intelligent robotics and autonomous agents. MIT Press, 2006.

\bibitem[Van~Horn et~al.(2015)Van~Horn, Branson, Farrell, Haber, Barry,
  Ipeirotis, Perona, and Belongie]{Horn_2015}
Grant Van~Horn, Steve Branson, Ryan Farrell, Scott Haber, Jessie Barry, Panos
  Ipeirotis, Pietro Perona, and Serge Belongie.
\newblock Building a bird recognition app and large scale dataset with citizen
  scientists: The fine print in fine-grained dataset collection.
\newblock In \emph{2015 IEEE Conference on Computer Vision and Pattern
  Recognition (CVPR)}, page 595–604. IEEE, Jun 2015.

\bibitem[Wang et~al.(2019)Wang, Ma, Chen, Luo, Yi, and
  Bailey]{Wang_Ma_Chen_Luo_Yi_Bailey_2019}
Yisen Wang, Xingjun Ma, Zaiyi Chen, Yuan Luo, Jinfeng Yi, and James Bailey.
\newblock Symmetric cross entropy for robust learning with noisy labels.
\newblock \emph{arXiv:1908.06112 [cs, stat]}, Aug 2019.

\bibitem[Xie et~al.(2016)Xie, Kiefel, Sun, and Geiger]{Xie2016CVPR}
Jun Xie, Martin Kiefel, Ming-Ting Sun, and Andreas Geiger.
\newblock Semantic instance annotation of street scenes by 3d to 2d label
  transfer.
\newblock In \emph{Conference on Computer Vision and Pattern Recognition
  (CVPR)}, 2016.

\bibitem[Xie et~al.(2020)Xie, Kiefel, Ming-Ting, and Gieger]{Kitti360}
Jun Xie, Martin Kiefel, Sun Ming-Ting, and Andreas Gieger.
\newblock Kitti-360, 2020.
\newblock URL \url{http://www.cvlibs.net/datasets/kitti-360/}.

\bibitem[Yi and Wu(2019)]{Yi_Wu_2019}
Kun Yi and Jianxin Wu.
\newblock Probabilistic end-to-end noise correction for learning with noisy
  labels.
\newblock \emph{arXiv:1903.07788 [cs]}, Mar 2019.

\bibitem[Yu et~al.(2019)Yu, Han, Yao, Niu, Tsang, and
  Sugiyama]{Yu_Han_Yao_Niu_Tsang_Sugiyama_2019}
Xingrui Yu, Bo~Han, Jiangchao Yao, Gang Niu, Ivor~W. Tsang, and Masashi
  Sugiyama.
\newblock How does disagreement help generalization against label corruption?
\newblock \emph{arXiv:1901.04215 [cs, stat]}, May 2019.

\bibitem[Zhang et~al.(2017)Zhang, Bengio, Hardt, Recht, and
  Vinyals]{Zhang_Bengio_Hardt_Recht_Vinyals_2017}
Chiyuan Zhang, Samy Bengio, Moritz Hardt, Benjamin Recht, and Oriol Vinyals.
\newblock Understanding deep learning requires rethinking generalization.
\newblock \emph{arXiv:1611.03530 [cs]}, Feb 2017.

\end{thebibliography}
